\pdfoutput=1

\documentclass[11pt]{article}
\usepackage{authblk}

\usepackage{emnlp2021}

\usepackage{times}
\usepackage{latexsym}

\usepackage[T1]{fontenc}

\usepackage[utf8]{inputenc}

\usepackage{microtype}

\usepackage{soul}
\usepackage{multirow}
\usepackage{tabularx}

\usepackage{multicol}
\usepackage{graphicx}
\usepackage{booktabs}

\title{Multilingual Counter Narrative Type Classification}

    \author[$\Diamond$, $\clubsuit$]{Yi-Ling Chung}
    \author[$\clubsuit$]{Marco Guerini}
    \author[$\spadesuit$]{Rodrigo Agerri}
    \affil[$\Diamond$]{University of Trento}
    \affil[$\clubsuit$]{Fondazione Bruno Kessler}
    \affil[$\spadesuit$]{HiTZ Center - Ixa, University of the Basque Country UPV/EHU \protect\\ 
    \protect\\ \texttt{ychung@fbk.eu, guerini@fbk.eu, rodrigo.agerri@ehu.eus}
     }
     
\begin{document}
\maketitle

\begin{abstract}
The growing interest in employing counter narratives for hatred intervention brings with it a focus on dataset creation and automation strategies. In this scenario, learning to recognize counter narrative types from natural text is expected to be useful for applications such as hate speech countering, where operators from non-governmental organizations are supposed to answer to hate with several and diverse arguments that can be mined from online sources. This paper presents the first multilingual work on counter narrative type classification, evaluating SoTA pre-trained language models in monolingual, multilingual and cross-lingual settings. When considering a fine-grained annotation of counter narrative classes, we report strong baseline classification results for the majority of the counter narrative types, especially if we translate every language to English before cross-lingual prediction. This suggests that knowledge about counter narratives can be successfully transferred across languages.

\end{abstract}

\section{Introduction}
The pervasive problem of online hate speech (HS) has motivated the research community to investigate methods for mitigating hatred, such as hate speech detection \cite{schmidt2017survey, fortuna2018survey} and, more recently, hate moderation through counter narratives (CNs). CNs are non-negative responses to hateful messages providing fact-bound arguments or alternative viewpoints. Distinct from standard approaches to hate intervention by content moderation, counter narratives are preferable as they preserve the right to freedom of speech and encourage peaceful conversations \cite{benesch2014countering, schieb2016governing}.

In particular, the NLP community has started exploring CN generation \cite{qian2019benchmark, tekiroglu2020generating, chung-etal-2021-towards,fanton-etal-2021-human, zhu-bhat-2021-generate}, also in multilingual settings \cite{chung2020italian} with the aim of helping non-governmental organizations (NGOs) fight HS.  However, automatic generation and detection of counter narratives still face three important challenges. First, counter narratives vary considerably in manner of expression and strategies \cite{benesch2016}, see Table ~\ref{tab:CN-example}, posing difficulties to automatic CN evaluation and CN classification. Second, few corpora are available since domain experts are required to obtain high-quality data. Third, the only existing work on classifying counter narratives targeted just English \cite{mathew2019thou}. 

In our view, automated mechanisms to distinguish CN characteristics are crucial for developing hate countering applications that can address personalization or for tasks such as counter narrative mining from user-generated content. Classifying counter narratives would also help to establish which counter narratives are more effective to target hate speech. Finally, it may also be useful for NGO operators to answer hate speech with diverse counter narratives (not just the most common ones) which could be mined from online sources.

\begin{table}[ht!]
\small
  \centering
  \begin{tabular}{p{0.95\linewidth}}
    \hline
\textbf{HS:} \textit{Muslims and non-British people are raping, enslaving and murdering our women! How disgusting!} \\  \hline
\textbf{CN1 (Denouncing):} \textit{This is not acceptable. Hatred cannot be tolerated and must be stopped.} \\  \hline
\textbf{CN2 (Facts):} \textit{Please notice that, regardless of religions, huge numbers of girls and children experience domestic sexual abuse by a partner or family member. We should encourage people to help victims and prevent sexual crimes.} \\ \hline
  \end{tabular}
 \caption{An example hate speech with counter narratives (CN1 and CN2) using different strategies.}
 \label{tab:CN-example}
\end{table}

To the best of our knowledge, this work is the first to tackle counter narrative classification from a multilingual perspective, predicting the types/strategies employed in a counter narrative message given a hateful message. Other contributions include: (1) the first empirical analysis of counter narrative characteristics on expert-based hate countering data targeting Islamophobia \cite{chung-etal-2019-conan}; (2) strong baseline results for multilingual and multi-class counter narrative classification by experimenting with the large pre-trained model XLM-R \cite{conneau-etal-2020-unsupervised}; and (3) a set of cross-lingual zero-shot experiments showing that huge improvements can be obtained by translating the training data from other languages to English, which is coherent with previous work on the multilinguality of language models \cite{pires-etal-2019-multilingual}. This result also shows that common characteristics of some counter narrative types are carried across languages. Data splits and training scripts are available at \textcolor{red}{\url{https://github.com/yilingchung/multilingualCN-classification}}.

\section{Counter Arguments v.s. Counter Narratives}
While counter arguments and counter narratives have a similar purpose  - to present an alternative stance to a statement - a counter argument is not necessarily a counter narrative, and vice versa. The main differences between the two are briefly discussed below, and point for the development of specific classification approaches for CNs. 

\paragraph{Presence of an argument.} A counter argument is typically defined as an argument (equivalent to a claim or standpoint) contradicting the initial statement with supporting evidence or reasoning \cite{khishfe2012relationship, Schiappa2013Argumentation,stab-etal-2018-cross}. The presence of supported evidence or reasoning is essential for counter-argumentation and argumentative information retrieval, regardless of it being implicit or explicit (e.g., Statement: \textit{Nuclear energy should be used to replace fossil fuels.} v.s. Counter argument: \textit{Renewable energy is a better option for replacing fossil fuels than unsustainable and expensive energy like nuclear power.}). 
In contrast, counter narratives can be formulated without using arguments. For example, a counter narrative can be without supporting evidence (e.g., Hate speech: \textit{Islam is a plague.} v.s. Counter narrative: \textit{How can you use such language to describe a religion of 1.6 billion people?}). 

\paragraph{Non-negative arguments.} Following guidelines\footnote{As an example: https://getthetrollsout.org/stoppinghate} generally adopted by NGOs, an appropriate counter narrative should not contain hostile language, prejudice, and unlawful content. Using negative tones as responses may risk escalating the conversation and engaging in hateful sentiment. On the contrary, counter arguments are not bound by such constraints.

\section{Counter Narrative Type Classification}
Classifying counter narrative types is a new and challenging task. We hypothesize that underlying linguistic features in one language can facilitate the classification performance in other languages. Accordingly, we approach this task in a multilingual setting focusing on English, French and Italian. We experimented in the following learning settings: (1) \textbf{monolingual}, in which the train and test set are in the same language; (2) \textbf{multilingual}, in which we train in 3 languages and test on each of them; (3) \textbf{zero-shot cross-lingual}, where we train in one language and evaluate on the other two unseen languages; and, (4) \textbf{zero-shot translated}, similar to the multilingual experiment but translating the Italian and French training data into English.

\subsection{Dataset}

\begin{table}[h]
\small 
\resizebox{\columnwidth}{!}{
\begin{tabular}{lrrr l rrr }
\hline
&\multicolumn{3}{c}{Train} & &\multicolumn{3}{c}{Test} \\\cline{2-4}\cline{6-8}
&               EN  &  IT  &  FR  &&  EN &  IT &  FR   \\ \cline{2-8}
Facts       &   957 & 1080 & 1329 && 237 & 270 & 333 \\
Question    &   258 &  285 &  342 &&  66 &  72 &  84 \\
Denouncing  &   405 &  153 &  705 && 102 &  36 & 174 \\
Humor       &   162 &   96 &  213 &&  42 &  24 &  51 \\  
Hypocrisy   &  270 &  111 &  348 &&  66 &  27 &  87 \\ 
\hline
Total       & 2052 & 1725 & 2937 && 513 & 429 & 729 \\
\hline
\end{tabular}
}
\caption{Number of instances by types.}
\label{tab:data_split}
\end{table}

In our experiments we use CONAN \cite{chung-etal-2019-conan}, the only multilingual hate countering dataset currently available that has been niche-sourced to NGO operators. The dataset consists of 14k HS-CN pairs with counter narrative type annotation over English, French, and Italian. For each language, an original pair is augmented with two paraphrases of the original HS coupled with the original CN. In CONAN annotation, one CN could be assigned more than one CN type. As a first investigation into multilingual counter narrative classification, we select the pairs annotated with just one CN type for simplicity, discarding 25\%, 17\%, and 27\% of samples for EN, FR and IT, respectively. Considering that the class balance is skewed, we narrow down the categories to majority classes over 3 languages: \textit{facts} (35\%), \textit{denouncing} (13\%), \textit{question} (9\%), \textit{hypocrisy} (7\%), and \textit{humor} (5\%). We then randomly sampled 80\% and 20\% of the dataset for the training and testing, ensuring that one HS and its two HS paraphrases paired with the same counter narrative are kept in the same split, so that the same CN does not appear by chance in both the training and testing sets. While the class imbalance poses a challenge, it reflects the practical scenario where certain types are less frequent.

Since CONAN contains only positive examples of CN, we further created 200 instances for the classes \textit{support} and \textit{unrelated}.\footnote{This allows, for instance, to address cases where we need to identify counter narratives in a conversation containing both abusive and non-abusive language.} Support is a response that endorses the hate speech and \textit{unrelated} is a text not connected to the hate speech in any sense. Clearly these two classes do not fall into the categories of CN and in fact they do not exist in CONAN. We include them to ensure that the models are exposed to varied non-CN text. Such setting can avoid model overfitting and increase the applicability of a system to the real world full of noisy content. The amount of these instances (200) is set to be close to the less populated class (\textit{humor}). For the \textit{unrelated} pairs, we randomly sampled data from Wikilingua\footnote{Wikilingua is a multilingual dataset for summarization covering how-to guides on various topics written by human from WikiHow.} \cite{ladhak-etal-2020-wikilingua}, featuring topics unrelated to islamophobia. The \textit{support} pairs consist of HS paired with each of the two paraphrases. Table \ref{tab:data_split} reports the distribution of training and testing examples per class across the three languages.

\subsection{Models}

The XLM-R language model \cite{conneau-etal-2020-unsupervised}, pre-trained on CommonCrawl for 100 languages, reports strong performance on several cross-lingual downstream tasks such as natural language inference or named entity recognition, and also on multilingual stance detection, close related to CN classification \cite{zotovaeswa21}. In this paper we leverage XLM-R to provide a strong baseline in the task of CN type classification in both monolingual and multilingual settings, using the Transformers library \cite{wolf-etal-2020-transformers}. For every experiment, we fine-tuned the base version of XLM-R for 10 epochs with batch size 32 and 2e-5 learning rate.

\section{Experimental Results}
Similarly to stance detection and argument mining tasks \cite{mohammad-etal-2016-semeval,stab-etal-2018-cross}, we report average macro F1 score over the CN types to avoid obtaining very high scores simply deriving from the dominant classes in the dataset. Furthermore, we also follow previous literature on stance detection \cite{mohammad-etal-2016-semeval} and report only the performance over the relevant classes.

\paragraph{Monolingual results.}

As shown by Table \ref{tab:experimental_result}, monolingual models consistently yield the best performance in predicting the majority classes \textit{facts} and \textit{question}. As for \textit{denouncing}, we obtain results above average for English, French and Italian (0.65, 0.58, and 0.51 respectively). Worst results are obtained for \emph{humor} and \emph{hypocrisy}. For Italian and French the model completely fails to identify \textit{humor}, the most difficult and under-represented class; for English, the prediction is moderate (0.45). Lastly, the low results for the \textit{hypocrisy} class seemed to be caused by the difficulty for the model in discriminating \emph{hypocrisy} from \textit{facts}. For example, 38\% of the prediction errors for the \emph{hypocrisy} class in English are caused by wrongly classifying \emph{hypocrisy} instances as \emph{facts} (more details are provided in the Appendix, Figure \ref{fig:confusion_metrics}).

In a post-hoc manual analysis, annotators expressed difficulties in differentiating these two classes, difficulties illustrated by the examples provided in Table \ref{tab:error_examples}. This issue seems to be further confirmed by annotation statistics obtained from CONAN: among all the instances that contains the   \textit{hypocrisy} label, 45\% were annotated with \textit{hypocrisy} alone, 27\% with \textit{hypocrisy} and \textit{facts}, while other multi-label cases were much lower. 
We hypothesize that \textit{hypocrisy} could be a subclass of \textit{facts}. After all, pointing out the contradiction in a hate statement may imply correcting misstatements via facts. Future work can try to merge two classes together to improve classification performance.

\begin{table*}[t]
\small
\centering
\begin{tabular}{lrrr l rrr l rrr l rrr}
\hline
&\multicolumn{3}{c}{Monolingual} && \multicolumn{3}{c}{Multilingual} &&\multicolumn{3}{c}{Zero-shot} &&\multicolumn{3}{c}{Zero-shot transl$_{en}$}\\
\cline{2-4}\cline{6-8}\cline{10-12}\cline{14-16}
Target language &  EN  &   IT &   FR &&   EN &   IT &  FR  &&  EN  &   IT &  FR  && EN &  IT &   FR\\ \cline{2-16}
Average       & 0.60 & 0.59 & 0.49 && \textbf{0.65} & 0.56 & 0.49 && 0.45 & 0.46 & 0.40 && \textbf{0.65} & \textbf{0.69} & \textbf{0.72}  \\
Facts         & 0.84 & 0.91 & 0.71 && 0.82 & 0.94 & 0.74 && 0.71 & 0.86 & 0.67 && \textbf{0.85} &  \textbf{0.95} & \textbf{0.83}\\
Question      & 0.73 & 0.91 & 0.68 && \textbf{0.81} & 0.91 & 0.69 && 0.65 & 0.85 & 0.52 && 0.75 &  \textbf{0.93} & \textbf{0.87} \\
Denouncing    & 0.65 & 0.51 & 0.58 && 0.68 & 0.46 & 0.61 && 0.28 & 0.36 & 0.39 && \textbf{0.70} & \textbf{0.69} & \textbf{0.75} \\
Humor         & 0.45 & 0.08 & 0.15 && 0.44 & 0.00 & 0.05 && 0.34 & 0.00 & 0.20 && \textbf{0.56} &  \textbf{0.42} & \textbf{0.51} \\
Hypocrisy     & 0.35 & 0.52 & 0.36 && \textbf{0.49} & \textbf{0.48} & 0.35 && 0.30 & 0.25 & 0.23 && 0.40 & 0.46 & \textbf{0.62}\\
\hline
\end{tabular}
\caption{Macro-F1 results under various configurations. For each target language in zero-shot setting, the scores are averaged over the models trained on the other two source languages.}
\label{tab:experimental_result}
\end{table*}

\begin{table*}[h!]
\small
  \centering
  \begin{tabularx}{\textwidth}{p{6.5cm}|X|l}
    \hline
HS & CN (Gold CN Type) & Prediction \\
    \hline 
\textit{Not all Muslims are terrorists, but all terrorists are Muslims.} & \textit{Only because when a non-muslim commits a terrorist act it is not called terrorism. Instead they're a `lone wolf' with a `traumatic past'.} (Hypocrisy) & Facts \\
    \hline 
\textit{Women of our culture that decide to become Islamic are so selfish. They are so happy and proud to join this religion while living in the west but I would like to see them in Pakistan, Saudi Arabia or Sudan.} & \textit{Maybe that `sisterhood of tolerance and understanding' is challenging sexism everywhere including those men who feel they can tell women what they should believe in and how to live their lives.} (Facts) & Hypocrisy \\
    \hline 
  \end{tabularx}
  \caption{Examples of misclassified counter narratives.}
 \label{tab:error_examples}
\end{table*}

\paragraph{Multilingual results.}

As it is reported in Table \ref{tab:experimental_result}, multilingual learning obtains better overall results than monolingual variants. In general the model trained on all languages improve the per-type performance although we observe decreases in \textit{denouncing} and \textit{hypocrisy} for Italian. Also, \textit{humor} is still challenging for each language with more than 50\% of miss-categorization. 

\paragraph{Zero-shot cross-lingual results.}
For each target language, we provide the average F1 of zero-shot cross-lingual models trained on the other 2 source languages in Table \ref{tab:experimental_result}. Although the results are lower than in the monolingual setting, results show that cross-lingual transfer can be a feasible strategy for classifying counter narratives for languages for which no training data is readily available, especially for the majority classes. Best results in this scenario are obtained when English is the source language, obtaining lower but close results with respect to the monolingual results in Italian and French (for more details about the results per language pair see Table \ref{tab:zeroshot_result_per_type} in the Appendix).

\paragraph{Zero-shot translated results.}

Data augmentation through translation has been widely employed to improve classification performance \cite{toledo-ronen-etal-2020-multilingual}, also in cross-lingual settings \cite{zotovaeswa21}. At the same time, the cross-lingual capabilities of Transformer models such as XLM-R and mBERT are being actively investigated \cite{muller-etal-2021-first,pires-etal-2019-multilingual,wu-dredze-2019-beto}. Thus, we conduct an additional experiment adding, to the English training set, the manually translated Italian and French training data before testing on the target languages (Italian and French). The aim is to investigate if XLM-R benefits from fine-tuning on a high resource language (English) instead of combining English with other languages, such as Italian and French, which are not so well represented in pre-trained multilingual models \cite{martin-etal-2020-camembert,agerri-etal-2020-give,Espinosa2020DeepReadingS}.  By doing so, in this `zero-shot transl$_{en}$' setting we aim to expose the model with semantic knowledge from some target languages without actually seeing those languages (Italian and French).
Table \ref{tab:experimental_result} shows that we obtain a huge performance jump with respect to the multilingual results, with 13 points improvement in macro-F1 score for Italian and 23 points for French. Interestingly, the improvement is more impressive for the \textit{humor} class, the most challenging of them all.

\section{Related Work}
Broadly speaking, counter narrative type classification is related to stance detection \cite{toledo-ronen-etal-2020-multilingual, schiller2021stance}, which is crucial for argument search \cite{stab-etal-2018-cross}. In contrast to stance detection, that concentrates on binary or relatively simple classification -- e.g., determine if an argument supports or contests a given topic \cite{sridhar-etal-2015-joint, rosenthal-mckeown-2015-couldnt, stab-etal-2018-cross} --  we present a multi-class approach to counter narrative classification.

Counter narratives have been adopted as a direct and effective response to online hatred in several campaigns and on social media platforms including Twitter \cite{munger2017tweetment, wright2017vectors}, Facebook \cite{schieb2016governing}, and Youtube \cite{ernst2017hate, mathew2019thou}. Although it has been argued that hate speech detection can benefit from CN classification, there are very few studies on this regard, with only one previous work on classifying counter narrative types \cite{mathew2019thou}. However, unlike our present work, they consider hostile language as one of the main types of counter narratives, which is explicitly discouraged by NGOs working on hatred intervention. Furthermore, we investigate multilingual and cross-lingual CN classification leveraging a SoTA pre-trained multilingual language model.  

\section{Conclusion}
We present the first work on multilingual CN type classification. Our results show that: (i) the performance is promising for the majority classes (\textit{facts}, \textit{question}, \textit{denouncing}); (ii) classifying \textit{humor} and \textit{hypocrisy} CNs is still challenging; (iii) combining training data from the three source languages improves performance over the monolingual evaluation; and (iv), the best overall results are obtained in the `zero-shot transl$_{en}$' approach where the training data for Italian and French is translated to English. This shows that some knowledge about CNs is transferred across languages. While this is coherent with previous literature about multilingual language models, the exact source of such successful transfer across languages remains an open topic.

\section*{Acknowledgements}

 Rodrigo Agerri is funded by the RYC-2017–23647 fellowship and by the DeepReading (RTI2018-096846-B-C21, MCIU/AEI/FEDER, UE, Spanish Ministry of Science, Innovation and Universities) and DeepText (KK-2020/00088, Basque Government) projects.

\bibliography{custom}

\begin{thebibliography}{34}
\expandafter\ifx\csname natexlab\endcsname\relax\def\natexlab#1{#1}\fi

\bibitem[{Agerri et~al.(2020)Agerri, San~Vicente, Campos, Barrena, Saralegi,
  Soroa, and Agirre}]{agerri-etal-2020-give}
Rodrigo Agerri, I{\~n}aki San~Vicente, Jon~Ander Campos, Ander Barrena, Xabier
  Saralegi, Aitor Soroa, and Eneko Agirre. 2020.
\newblock \href {https://www.aclweb.org/anthology/2020.lrec-1.588} {Give your
  text representation models some love: the case for {B}asque}.
\newblock In \emph{Proceedings of the 12th Language Resources and Evaluation
  Conference}, pages 4781--4788, Marseille, France. European Language Resources
  Association.

\bibitem[{Benesch(2014)}]{benesch2014countering}
Susan Benesch. 2014.
\newblock \href
  {https://www.ushmm.org/m/pdfs/20140212-benesch-countering-dangerous-speech.pdf}
  {Countering dangerous speech: New ideas for genocide prevention}.
\newblock \emph{Washington, DC: United States Holocaust Memorial Museum}.

\bibitem[{Benesch et~al.(2016)Benesch, Ruths, Dillon, Saleem, and
  Wright}]{benesch2016}
Susan Benesch, Derek Ruths, Kelly~P Dillon, Haji~Mohammad Saleem, and Lucas
  Wright. 2016.
\newblock \href
  {https://dangerousspeech.org/counterspeech-on-twitter-a-field-study/}
  {Counterspeech on twitter: A field study}.
\newblock \emph{Dangerous Speech Project}.

\bibitem[{Chung et~al.(2019)Chung, Kuzmenko, Tekiroglu, and
  Guerini}]{chung-etal-2019-conan}
Yi-Ling Chung, Elizaveta Kuzmenko, Serra~Sinem Tekiroglu, and Marco Guerini.
  2019.
\newblock \href {https://doi.org/10.18653/v1/P19-1271} {{CONAN} - {CO}unter
  {NA}rratives through nichesourcing: a multilingual dataset of responses to
  fight online hate speech}.
\newblock In \emph{Proceedings of the 57th Annual Meeting of the Association
  for Computational Linguistics}, pages 2819--2829, Florence, Italy.
  Association for Computational Linguistics.

\bibitem[{Chung et~al.(2020)Chung, Tekiroglu, and Guerini}]{chung2020italian}
Yi-Ling Chung, Serra~Sinem Tekiroglu, and Marco Guerini. 2020.
\newblock \href {http://ceur-ws.org/Vol-2769/paper_35.pdf} {Italian counter
  narrative generation to fight online hate speech}.
\newblock In \emph{Proceedings of the Seventh Italian Conference on
  Computational Linguistics}, Bologna, Italy.

\bibitem[{Chung et~al.(2021)Chung, Tekiro{\u{g}}lu, and
  Guerini}]{chung-etal-2021-towards}
Yi-Ling Chung, Serra~Sinem Tekiro{\u{g}}lu, and Marco Guerini. 2021.
\newblock \href {https://doi.org/10.18653/v1/2021.findings-acl.79} {Towards
  knowledge-grounded counter narrative generation for hate speech}.
\newblock In \emph{Findings of the Association for Computational Linguistics:
  ACL-IJCNLP 2021}, pages 899--914, Online. Association for Computational
  Linguistics.

\bibitem[{Conneau et~al.(2020)Conneau, Khandelwal, Goyal, Chaudhary, Wenzek,
  Guzm{\'a}n, Grave, Ott, Zettlemoyer, and
  Stoyanov}]{conneau-etal-2020-unsupervised}
Alexis Conneau, Kartikay Khandelwal, Naman Goyal, Vishrav Chaudhary, Guillaume
  Wenzek, Francisco Guzm{\'a}n, Edouard Grave, Myle Ott, Luke Zettlemoyer, and
  Veselin Stoyanov. 2020.
\newblock \href {https://doi.org/10.18653/v1/2020.acl-main.747} {Unsupervised
  cross-lingual representation learning at scale}.
\newblock In \emph{Proceedings of the 58th Annual Meeting of the Association
  for Computational Linguistics}, pages 8440--8451, Online. Association for
  Computational Linguistics.

\bibitem[{Ernst et~al.(2017)Ernst, Schmitt, Rieger, Beier, Vorderer, Bente, and
  Roth}]{ernst2017hate}
Julian Ernst, Josephine~B Schmitt, Diana Rieger, Ann~Kristin Beier, Peter
  Vorderer, Gary Bente, and Hans-Joachim Roth. 2017.
\newblock \href {https://journals.sfu.ca/jd/index.php/jd/article/view/91} {Hate
  beneath the counter speech? a qualitative content analysis of user comments
  on youtube related to counter speech videos}.
\newblock \emph{Journal for Deradicalization}, 10:1--49.

\bibitem[{Espinosa et~al.(2020)Espinosa, Agerri, Rodrigo, and
  Centeno}]{Espinosa2020DeepReadingS}
Mar{\'i}a~S. Espinosa, Rodrigo Agerri, {\'A}lvaro Rodrigo, and Roberto Centeno.
  2020.
\newblock Deepreading @ sardistance 2020: Combining textual, social and
  emotional features.
\newblock In \emph{EVALITA}.

\bibitem[{Fanton et~al.(2021)Fanton, Bonaldi, Tekiro{\u{g}}lu, and
  Guerini}]{fanton-etal-2021-human}
Margherita Fanton, Helena Bonaldi, Serra~Sinem Tekiro{\u{g}}lu, and Marco
  Guerini. 2021.
\newblock \href {https://doi.org/10.18653/v1/2021.acl-long.250}
  {Human-in-the-loop for data collection: a multi-target counter narrative
  dataset to fight online hate speech}.
\newblock In \emph{Proceedings of the 59th Annual Meeting of the Association
  for Computational Linguistics and the 11th International Joint Conference on
  Natural Language Processing (Volume 1: Long Papers)}, pages 3226--3240,
  Online. Association for Computational Linguistics.

\bibitem[{Fortuna and Nunes(2018)}]{fortuna2018survey}
Paula Fortuna and S{\'e}rgio Nunes. 2018.
\newblock A survey on automatic detection of hate speech in text.
\newblock \emph{ACM Computing Surveys (CSUR)}, 51(4):85.

\bibitem[{Khishfe(2012)}]{khishfe2012relationship}
Rola Khishfe. 2012.
\newblock Relationship between nature of science understandings and
  argumentation skills: A role for counterargument and contextual factors.
\newblock \emph{Journal of Research in Science Teaching}, 49(4):489--514.

\bibitem[{Ladhak et~al.(2020)Ladhak, Durmus, Cardie, and
  McKeown}]{ladhak-etal-2020-wikilingua}
Faisal Ladhak, Esin Durmus, Claire Cardie, and Kathleen McKeown. 2020.
\newblock \href {https://doi.org/10.18653/v1/2020.findings-emnlp.360}
  {{W}iki{L}ingua: A new benchmark dataset for cross-lingual abstractive
  summarization}.
\newblock In \emph{Findings of the Association for Computational Linguistics:
  EMNLP 2020}, pages 4034--4048, Online. Association for Computational
  Linguistics.

\bibitem[{Martin et~al.(2020)Martin, Muller, Ortiz~Su{\'a}rez, Dupont, Romary,
  de~la Clergerie, Seddah, and Sagot}]{martin-etal-2020-camembert}
Louis Martin, Benjamin Muller, Pedro~Javier Ortiz~Su{\'a}rez, Yoann Dupont,
  Laurent Romary, {\'E}ric de~la Clergerie, Djam{\'e} Seddah, and Beno{\^\i}t
  Sagot. 2020.
\newblock \href {https://doi.org/10.18653/v1/2020.acl-main.645} {{C}amem{BERT}:
  a tasty {F}rench language model}.
\newblock In \emph{Proceedings of the 58th Annual Meeting of the Association
  for Computational Linguistics}, pages 7203--7219, Online. Association for
  Computational Linguistics.

\bibitem[{Mathew et~al.(2019)Mathew, Saha, Tharad, Rajgaria, Singhania, Maity,
  Goyal, and Mukherjee}]{mathew2019thou}
Binny Mathew, Punyajoy Saha, Hardik Tharad, Subham Rajgaria, Prajwal Singhania,
  Suman~Kalyan Maity, Pawan Goyal, and Animesh Mukherjee. 2019.
\newblock Thou shalt not hate: Countering online hate speech.
\newblock In \emph{Proceedings of the International AAAI Conference on Web and
  Social Media}, volume~13, pages 369--380.

\bibitem[{Mohammad et~al.(2016)Mohammad, Kiritchenko, Sobhani, Zhu, and
  Cherry}]{mohammad-etal-2016-semeval}
Saif Mohammad, Svetlana Kiritchenko, Parinaz Sobhani, Xiaodan Zhu, and Colin
  Cherry. 2016.
\newblock \href {https://doi.org/10.18653/v1/S16-1003} {{S}em{E}val-2016 task
  6: Detecting stance in tweets}.
\newblock In \emph{Proceedings of the 10th International Workshop on Semantic
  Evaluation ({S}em{E}val-2016)}, pages 31--41, San Diego, California.
  Association for Computational Linguistics.

\bibitem[{Muller et~al.(2021)Muller, Elazar, Sagot, and
  Seddah}]{muller-etal-2021-first}
Benjamin Muller, Yanai Elazar, Beno{\^\i}t Sagot, and Djam{\'e} Seddah. 2021.
\newblock \href {https://www.aclweb.org/anthology/2021.eacl-main.189} {First
  align, then predict: Understanding the cross-lingual ability of multilingual
  {BERT}}.
\newblock In \emph{Proceedings of the 16th Conference of the European Chapter
  of the Association for Computational Linguistics: Main Volume}, pages
  2214--2231, Online. Association for Computational Linguistics.

\bibitem[{Munger(2017)}]{munger2017tweetment}
Kevin Munger. 2017.
\newblock \href {https://link.springer.com/article/10.1007/s11109-016-9373-5}
  {Tweetment effects on the tweeted: Experimentally reducing racist
  harassment}.
\newblock \emph{Political Behavior}, 39(3):629--649.

\bibitem[{Pires et~al.(2019)Pires, Schlinger, and
  Garrette}]{pires-etal-2019-multilingual}
Telmo Pires, Eva Schlinger, and Dan Garrette. 2019.
\newblock \href {https://doi.org/10.18653/v1/P19-1493} {How multilingual is
  multilingual {BERT}?}
\newblock In \emph{Proceedings of the 57th Annual Meeting of the Association
  for Computational Linguistics}, pages 4996--5001, Florence, Italy.
  Association for Computational Linguistics.

\bibitem[{Qian et~al.(2019)Qian, Bethke, Liu, Belding, and
  Wang}]{qian2019benchmark}
Jing Qian, Anna Bethke, Yinyin Liu, Elizabeth Belding, and William~Yang Wang.
  2019.
\newblock \href {https://doi.org/10.18653/v1/D19-1482} {A benchmark dataset for
  learning to intervene in online hate speech}.
\newblock In \emph{Proceedings of the 2019 Conference on Empirical Methods in
  Natural Language Processing and the 9th International Joint Conference on
  Natural Language Processing}, pages 4755--4764, Hong Kong, China. Association
  for Computational Linguistics.

\bibitem[{Rosenthal and McKeown(2015)}]{rosenthal-mckeown-2015-couldnt}
Sara Rosenthal and Kathy McKeown. 2015.
\newblock \href {https://doi.org/10.18653/v1/W15-4625} {{I} couldn{'}t agree
  more: The role of conversational structure in agreement and disagreement
  detection in online discussions}.
\newblock In \emph{Proceedings of the 16th Annual Meeting of the Special
  Interest Group on Discourse and Dialogue}, pages 168--177, Prague, Czech
  Republic. Association for Computational Linguistics.

\bibitem[{Schiappa and Nordin(2013)}]{Schiappa2013Argumentation}
Edward Schiappa and John~P. Nordin. 2013.
\newblock \emph{Argumentation: Keeping Faith with ReasonKeeping Faith With
  Reason}.
\newblock Pearson UK.

\bibitem[{Schieb and Preuss(2016)}]{schieb2016governing}
Carla Schieb and Mike Preuss. 2016.
\newblock Governing hate speech by means of counterspeech on facebook.
\newblock In \emph{66th ICA Annual Conference}, pages 1--23, Fukuoka, Japan.

\bibitem[{Schiller et~al.(2021)Schiller, Daxenberger, and
  Gurevych}]{schiller2021stance}
Benjamin Schiller, Johannes Daxenberger, and Iryna Gurevych. 2021.
\newblock Stance detection benchmark: How robust is your stance detection?
\newblock \emph{KI-K{\"u}nstliche Intelligenz}, pages 1--13.

\bibitem[{Schmidt and Wiegand(2017)}]{schmidt2017survey}
Anna Schmidt and Michael Wiegand. 2017.
\newblock A survey on hate speech detection using natural language processing.
\newblock In \emph{Proceedings of the Fifth International Workshop on Natural
  Language Processing for Social Media}, pages 1--10.

\bibitem[{Sridhar et~al.(2015)Sridhar, Foulds, Huang, Getoor, and
  Walker}]{sridhar-etal-2015-joint}
Dhanya Sridhar, James Foulds, Bert Huang, Lise Getoor, and Marilyn Walker.
  2015.
\newblock \href {https://doi.org/10.3115/v1/P15-1012} {Joint models of
  disagreement and stance in online debate}.
\newblock In \emph{Proceedings of the 53rd Annual Meeting of the Association
  for Computational Linguistics and the 7th International Joint Conference on
  Natural Language Processing (Volume 1: Long Papers)}, pages 116--125,
  Beijing, China. Association for Computational Linguistics.

\bibitem[{Stab et~al.(2018)Stab, Miller, Schiller, Rai, and
  Gurevych}]{stab-etal-2018-cross}
Christian Stab, Tristan Miller, Benjamin Schiller, Pranav Rai, and Iryna
  Gurevych. 2018.
\newblock \href {https://doi.org/10.18653/v1/D18-1402} {Cross-topic argument
  mining from heterogeneous sources}.
\newblock In \emph{Proceedings of the 2018 Conference on Empirical Methods in
  Natural Language Processing}, pages 3664--3674, Brussels, Belgium.
  Association for Computational Linguistics.

\bibitem[{Tekiro{\u{g}}lu et~al.(2020)Tekiro{\u{g}}lu, Chung, and
  Guerini}]{tekiroglu2020generating}
Serra~Sinem Tekiro{\u{g}}lu, Yi-Ling Chung, and Marco Guerini. 2020.
\newblock \href {https://doi.org/10.18653/v1/2020.acl-main.110} {Generating
  counter narratives against online hate speech: Data and strategies}.
\newblock In \emph{Proceedings of the 58th Annual Meeting of the Association
  for Computational Linguistics}, pages 1177--1190, Online. Association for
  Computational Linguistics.

\bibitem[{Toledo-Ronen et~al.(2020)Toledo-Ronen, Orbach, Bilu, Spector, and
  Slonim}]{toledo-ronen-etal-2020-multilingual}
Orith Toledo-Ronen, Matan Orbach, Yonatan Bilu, Artem Spector, and Noam Slonim.
  2020.
\newblock \href {https://doi.org/10.18653/v1/2020.findings-emnlp.29}
  {Multilingual argument mining: Datasets and analysis}.
\newblock In \emph{Findings of the Association for Computational Linguistics:
  EMNLP 2020}, pages 303--317, Online. Association for Computational
  Linguistics.

\bibitem[{Wolf et~al.(2020)Wolf, Debut, Sanh, Chaumond, Delangue, Moi, Cistac,
  Rault, Louf, Funtowicz, Davison, Shleifer, von Platen, Ma, Jernite, Plu, Xu,
  Le~Scao, Gugger, Drame, Lhoest, and Rush}]{wolf-etal-2020-transformers}
Thomas Wolf, Lysandre Debut, Victor Sanh, Julien Chaumond, Clement Delangue,
  Anthony Moi, Pierric Cistac, Tim Rault, Remi Louf, Morgan Funtowicz, Joe
  Davison, Sam Shleifer, Patrick von Platen, Clara Ma, Yacine Jernite, Julien
  Plu, Canwen Xu, Teven Le~Scao, Sylvain Gugger, Mariama Drame, Quentin Lhoest,
  and Alexander Rush. 2020.
\newblock \href {https://doi.org/10.18653/v1/2020.emnlp-demos.6} {Transformers:
  State-of-the-art natural language processing}.
\newblock In \emph{Proceedings of the 2020 Conference on Empirical Methods in
  Natural Language Processing: System Demonstrations}, pages 38--45, Online.
  Association for Computational Linguistics.

\bibitem[{Wright et~al.(2017)Wright, Ruths, Dillon, Saleem, and
  Benesch}]{wright2017vectors}
Lucas Wright, Derek Ruths, Kelly~P Dillon, Haji~Mohammad Saleem, and Susan
  Benesch. 2017.
\newblock \href {https://doi.org/10.18653/v1/W17-3009} {Vectors for
  counterspeech on {T}witter}.
\newblock In \emph{Proceedings of the First Workshop on Abusive Language
  Online}, pages 57--62, Vancouver, BC, Canada. Association for Computational
  Linguistics.

\bibitem[{Wu and Dredze(2019)}]{wu-dredze-2019-beto}
Shijie Wu and Mark Dredze. 2019.
\newblock \href {https://doi.org/10.18653/v1/D19-1077} {Beto, bentz, becas: The
  surprising cross-lingual effectiveness of {BERT}}.
\newblock In \emph{Proceedings of the 2019 Conference on Empirical Methods in
  Natural Language Processing and the 9th International Joint Conference on
  Natural Language Processing (EMNLP-IJCNLP)}, pages 833--844, Hong Kong,
  China. Association for Computational Linguistics.

\bibitem[{Zhu and Bhat(2021)}]{zhu-bhat-2021-generate}
Wanzheng Zhu and Suma Bhat. 2021.
\newblock \href {https://doi.org/10.18653/v1/2021.findings-acl.12} {Generate,
  prune, select: A pipeline for counterspeech generation against online hate
  speech}.
\newblock In \emph{Findings of the Association for Computational Linguistics:
  ACL-IJCNLP 2021}, pages 134--149, Online. Association for Computational
  Linguistics.

\bibitem[{Zotova et~al.(2021)Zotova, Agerri, and Rigau}]{zotovaeswa21}
Elena Zotova, Rodrigo Agerri, and German Rigau. 2021.
\newblock \href {https://doi.org/10.1016/j.eswa.2020.114547} {Semi-automatic
  generation of multilingual datasets for stance detection in twitter}.
\newblock \emph{Expert Syst. Appl.}, 170:114547.

\end{thebibliography}
\bibliographystyle{acl_natbib}
\appendix
\clearpage
\onecolumn
\section{Appendices}

\begin{table*}[ht!]
\centering
\begin{tabular}{lrrr | rrr|rr }
              &  IT -> EN  & FR -> EN && EN -> IT & FR -> IT && EN -> FR & IT -> FR   \\ \hline
Average       &     0.41   &   0.50   &&    0.48  &     0.45 &&     0.43 &      0.37  \\
Facts         &     0.71   &   0.71   &&    0.88  &     0.84 &&     0.73 &      0.60  \\
Question      &     0.56   &   0.74  &&    0.84  &     0.87 &&     0.48 &      0.55  \\
Denouncing    &     0.30   &   0.25   &&    0.46  &     0.26 &&     0.44 &      0.33  \\
Humor         &     0.24   &   0.43   &&    0.00  &     0.00 &&     0.22 &      0.18  \\
Hypocrisy     &     0.22   &   0.38   &&    0.24  &     0.27 &&     0.28 &      0.17  \\
\hline
\end{tabular}
\caption{Zero-shot cross-lingual results in terms of macro-F1 per type.}
\label{tab:zeroshot_result_per_type}
\end{table*}

\label{sec:appendix}
\begin{figure*}[ht!]
\begin{multicols}{3}
    \includegraphics[width=\linewidth]{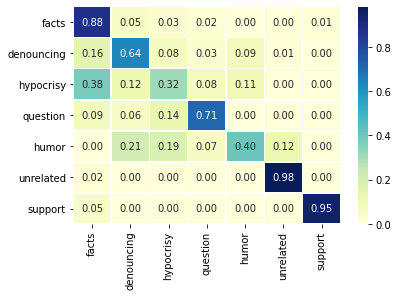}\par 
    \includegraphics[width=\linewidth]{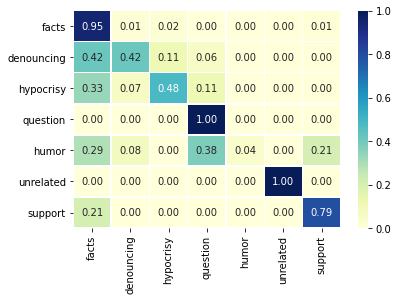}\par 
    \includegraphics[width=\linewidth]{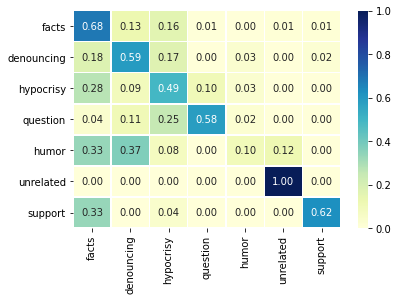}\par 
    \end{multicols}
\begin{multicols}{3}
    \includegraphics[width=\linewidth]{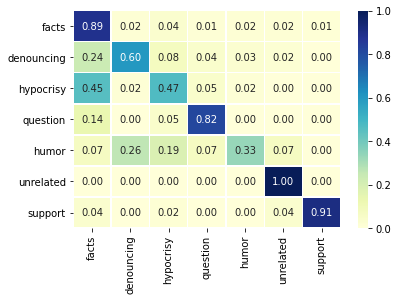}\par
    \includegraphics[width=\linewidth]{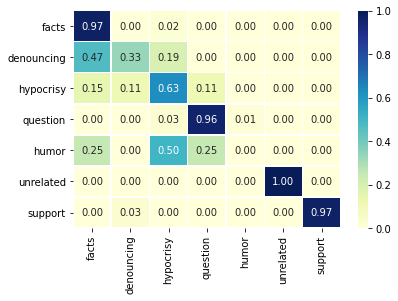}\par
    \includegraphics[width=\linewidth]{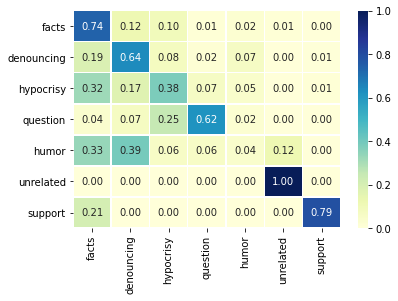}\par
\end{multicols}
\caption{Confusion matrix on monolingual training for EN, IT, and FR from left to right (upper part); multilingual model tested on EN, IT, and FR from left to right (down part). The predictions are represented by columns and gold class is represented in rows.}
\label{fig:confusion_metrics}
\end{figure*}

\end{document}